\documentclass[final]{cvpr}

\usepackage{times}
\usepackage{epsfig}
\usepackage{graphicx}
\usepackage{amsmath}
\usepackage{amssymb}

\usepackage{url}
\usepackage{xspace}
\usepackage{color}
\usepackage{array}
\usepackage{subfigure}

\graphicspath{{images/}}

\usepackage[pagebackref=true,breaklinks=true,colorlinks,bookmarks=false]{hyperref}

\def\cvprPaperID{12} 
\def\confYear{EarthVision 2021}

\newcommand{\datasetname}{LandCover.ai\xspace}
\newcommand{\datasetlongname}{LandCover.ai (\textbf{Land Cover} from \textbf{A}erial \textbf{I}magery)\xspace}
\newcommand{\dataseturl}{\url{https://landcover.ai.linuxpolska.com/}\xspace}

\begin{document}

\title{\datasetname: Dataset for Automatic Mapping of Buildings, Woodlands, Water and Roads from Aerial Imagery}

\author{Adrian Boguszewski\\
Linux Polska\\
Warsaw, Poland\\
{\tt\small adrian.boguszewski@linuxpolska.pl}
\and
Dominik Batorski\\
Interdisciplinary Centre for Mathematical and Computational Modelling\\
University of Warsaw, Poland\\
{\tt\small batorski@uw.edu.pl}
\and
Natalia Ziemba-Jankowska\\
Linux Polska\\
Warsaw, Poland\\
{\tt\small natalia.ziemba-jankowska@linuxpolska.pl}
\and
Tomasz Dziedzic\\
Linux Polska\\
Warsaw, Poland\\
{\tt\small tomasz.dziedzic@linuxpolska.pl}
\and
Anna Zambrzycka\\
Agency for Restructuring and Modernisation of Agriculture\\
Warsaw, Poland\\
{\tt\small anna.zambrzycka@arimr.gov.pl}
}

\maketitle

\begin{abstract}
Monitoring of land cover and land use is crucial in natural resources management. Automatic visual mapping can carry enormous economic value for agriculture, forestry, or public administration. Satellite or aerial images combined with computer vision and deep learning enable precise assessment and can significantly speed up change detection. Aerial imagery usually provides images with much higher pixel resolution than satellite data allowing more detailed mapping. However, there is still a lack of aerial datasets made for the segmentation, covering rural areas with a resolution of tens centimeters per pixel, manual fine labels, and highly publicly important environmental instances like buildings, woods, water, or roads. 

Here we introduce \datasetlongname dataset for semantic segmentation. We collected images of 216.27~km\textsuperscript{2} rural areas across Poland, a country in Central Europe, 39.51~km\textsuperscript{2} with resolution 50~cm per pixel and 176.76~km\textsuperscript{2} with resolution 25~cm per pixel and manually fine annotated four following classes of objects: buildings, woodlands, water, and roads. Additionally, we report simple benchmark results, achieving 85.56\% of mean intersection over union on the test set. It proves that the automatic mapping of land cover is possible with a relatively small, cost-efficient, RGB-only dataset. The dataset is publicly available at \dataseturl
\end{abstract}

\section{Introduction}

Monitoring and assessment of land cover and land use are essential in natural resources management. Remote sensing data and image processing techniques have been widely used to provide a land description and change detection in urban and countryside areas. Detailed information about land use or land cover is a valuable source in various fields, such as urban planning \cite{Pauleit2000,Zhou2011}, change detection \cite{Gerard2010}, vegetation monitoring \cite{Ahmed2017}, or even military reconnaissance. Changes in land cover are important as an indicator of environmental change \cite{Wickham2000,Weber2001}, forest cover dynamics \cite{Potapov2015}, and degradation \cite{Kennedy2010} as well as one of the methods of biodiversity monitoring \cite{Pereira2006}. This type of data can be used to investigate processes that take place in the landscape, such as flows between various land covers \cite{Feranec2010} allowing to study the rate of urbanization, deforestation, agricultural intensity, and other man-made changes.

The majority of those studies use multispectral satellite imagery. Though such data are useful for many purposes, the resolution of available free satellite data is generally between 10 and 30~m \cite{Wulder2012} and high-resolution commercial satellite images are rather expensive. The aerial photographs, often done by local and state government, usually have a pixel size of 25-50~cm or even lower.

Aerial imagery used as a proper assessment of land parcel content carries high economic values for agricultural and public administration. Taxes and government subsidies depend on the nature of the parcel usage \eg plantations or buildings. However, human activity results in frequent changes in the appearance and usage of parcels. To discover these changes, new aerial orthophotos are required. Then it is necessary to indicate and vectorize all objects that have emerged or disappeared. For example, public agencies paying subsidies need to detect changes in classes of objects not eligible for the payments. From the business point of view, buildings, trees (including forests), water and roads are the most common objects. These classes are also affected by the dynamics of change over time.

Typically, the change detection process is carried out manually or using simple image classification (object or pixel \cite{Blaschke2010,Khatami2016,Caccetta2016,Ahmed2017}). Most GIS-based programs provide a large number of such tools. The operator browses individual orthophotos images and physically indicates the "new" objects. The whole process usually takes months throughout the country. Our own experience shows that for the area of over 312~000~km\textsuperscript{2} this workflow is very time-consuming and results in a significant error level (on average above 30\% over multiple years). Consequently, it is expensive to manually encode large areas of land \cite{Gerard2010}. Therefore, it becomes necessary to develop an efficient automatic tool to shorten the processing time and ensure higher accuracy. 

The classical computer vision approach, based on manually crafted feature extractors and rules, is insufficient in case of high variance and large scale of data and effects in the high effort and poor scalability. Deep learning with convolutional neural networks (CNN) has started to play a critical role in automatic change detection on aerial images \cite{CNNCV,7592858,MnihThesis,semlabCNN}. The unique composition of features such as scalability, affordability, and performance allows for quick and accurate monitoring of much larger regions. Moreover, it provides the possibility to detect changes over time \eg by comparing the semantic segmentation of particular areas for different moments. Unfortunately, the deep learning approach usually requires large datasets with ground truth annotations. 

While aerial images are readily obtainable, the efforts to generate high-quality datasets are limited by the enormous effort required to create accompanying annotations. Similar to other domains, the lack of natural annotated datasets is a limiting factor in the use of computer vision to land cover segmentation. A few fine-annotated datasets have been released recently \cite{isprs,chiu2020agriculturevision,Chen_2019,maggiori2017dataset}. On the other hand, \cite{Kaiser2017} uses RGB orthophotos from \emph{Google Maps} together with weakly labeled training data automatically derive from \emph{OpenStreetMap} to detect buildings and roads.

However, none of the above provide segmentation of buildings, woodlands, water, and roads simultaneously. To address this issue, we introduce \datasetlongname dataset suitable for semantic segmentation, which contains four manually annotated classes mentioned above. We collected images of 216.27~km\textsuperscript{2} of lands across Poland, a medium-sized country in Central Europe, 39.51~km\textsuperscript{2} with resolution 50~cm per pixel and 176.76~km\textsuperscript{2} with resolution 25~cm per pixel. Furthermore, we provide some results of a baseline model as a benchmark for comparison.

\section{Related works}

As mentioned, deep convolutional neural networks offer significant speedup over the previous manual work but require properly annotated data. Most segmentation datasets focus primarily on common objects or street views \cite{Cordts_2016_CVPR,1405.0312}, but aerial or satellite imagery requires a different perspective and an adequate set of classes. There are some datasets with aerial and satellite images. 

One of the earliest satellite datasets is UC-Merced \cite{inproceedings} with 30 cm per pixel resolution and 21 categories like buildings, forest, and even rivers. However, it is prepared for a classification of whole images, which is insufficient for the segmentation task. Other similar datasets like WHU-RS Dataset \cite{WHURS}, RSSCN7 \cite{RSSCN7}, AID \cite{1608.05167}, NWPU-RESISC45 \cite{NWPU-RESISC45}, and PatternNet \cite{1706.03424}, which are mostly collected from \emph{Google Maps}, are for image classification also. 

On the other hand, DOTA \cite{1807.02700} and iSAID \cite{1905.12886} are aerial datasets made for multi-class detection and instance segmentation respectively. Even though they have many categories (15 classes like vehicles, bridges, ships, but also courts and game fields), they are inadequate for public agencies' responsibilities like natural resources management. 

Datasets created only for buildings or vegetation segmentation are useful but not entirely sufficient in land cover change detection. Those datasets include the Massachusetts Buildings Dataset \cite{MnihThesis}, the Inria Aerial Image Labeling Dataset \cite{maggiori2017dataset}, the AIRS Automatic Mapping of Buildings Dataset \cite{Chen_2019}, the Agriculture-Vision a Large Aerial Image Database for Agricultural Pattern Analysis \cite{chiu2020agriculturevision}, and the Tree Cover dataset for the year 2010 of the Metropolitan Region of São Paulo \cite{treecover}. 

ISPRS Vaihingen and Potsdam datasets \cite{isprs} are manually annotated, but they are relatively small (fewer than 5 km\textsuperscript{2} of labeled data). Although they have the most common classes, water class is missing. Moreover, they mainly cover urban areas.

Chesapeake Bay Land Cover Dataset \cite{robinson2019large} is very large ($\sim$160,000 km\textsuperscript{2}, 2\% of the USA), but has a lower resolution (1 m) than desired and seems to be automatically annotated. There are many useful classes (\eg water, tree canopy, impervious roads) but the category "building" is missing.

Some popular aerial datasets are made with unmanned aerial vehicles (UAVs) imagery. There are even UAVs video datasets like ERA \cite{mou2020era}, UCLA Aerial Eve \cite{1706.03038}, or Okutama-Action \cite{1505.05957}, but made for event recognition, therefore they are not applicable.

\section{The dataset}

We decided to create a simple RGB-only dataset, fully manually annotated, large, and diverse enough to train a model for accurate semantic segmentation.

\begin{figure}[t]
\centering
\includegraphics[width=0.8\linewidth]{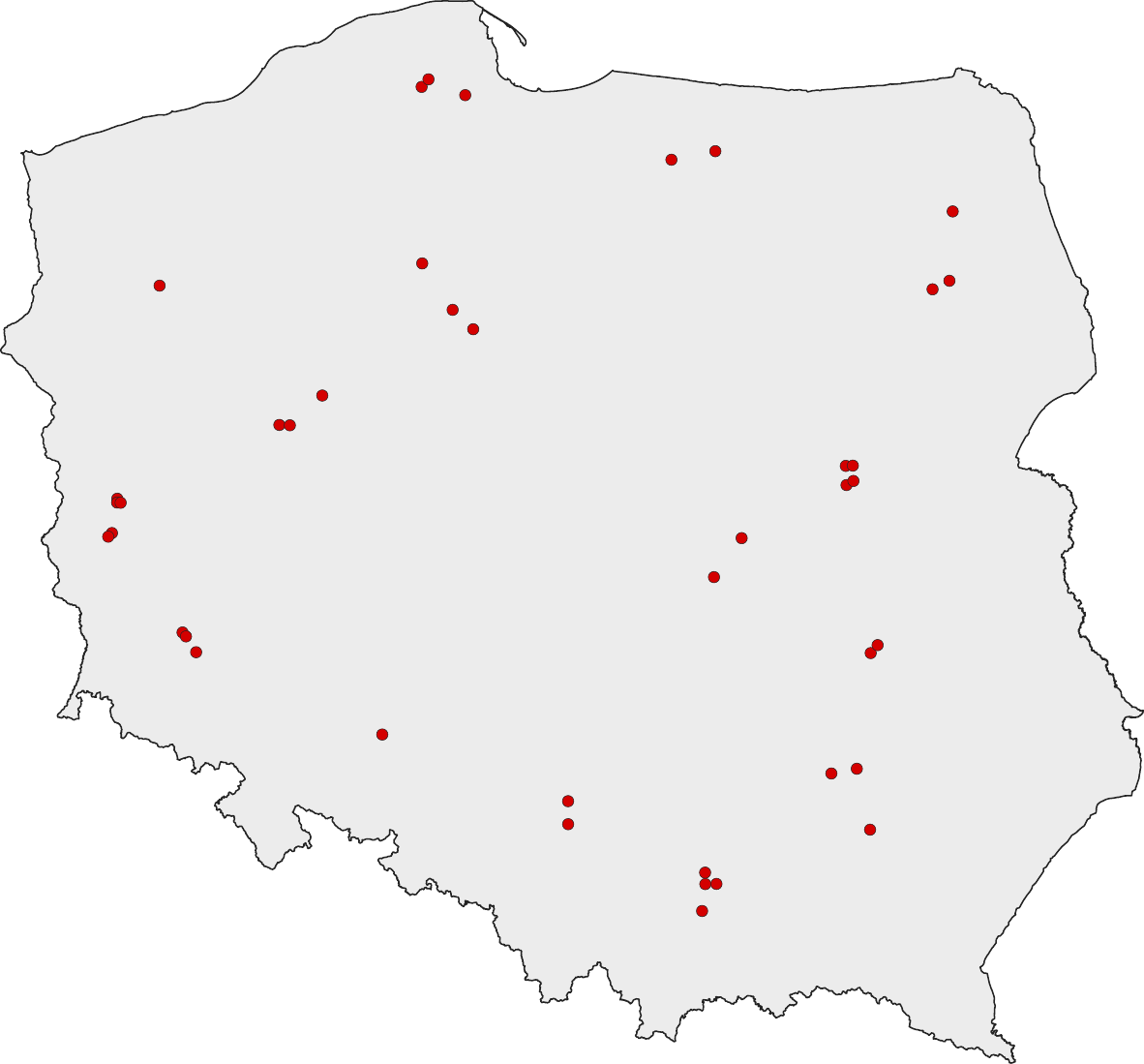}
\caption{Locations of selected orthophotos. The images were taken from areas of diverse morphological, agrarian, and vegetation conditions.}
\label{fig:poland}
\end{figure}

\subsection{Data acquisition}

The dataset consists of images selected from aerial photos used to develop the digital orthophoto covering the whole of Poland. All images come from the public geodetic resource and are compiled to update reference data of the land parcel identification system (LPIS). Digital orthophotos are made in cartesian "1992" (EPSG:2180) co-ordinate spatial reference system. Pictures were taken in spatial resolution of 25 or 50~cm per pixel with three spectral bands RGB. They come from different years (2015 - 2018) and flights. The photo-flying season in Poland begins in April and lasts until the end of September. Therefore, the acquired photos are characterized by a wide variety of optical conditions. They include images of different saturation, angles of sunlight, and shadow lengths. Simultaneously, the photos are from varying periods of the vegetation season. It makes this dataset robust and more applicable.

\begin{figure}[t]
    \centering
    \subfigure{\includegraphics[width=0.49\linewidth, height=0.49\linewidth]{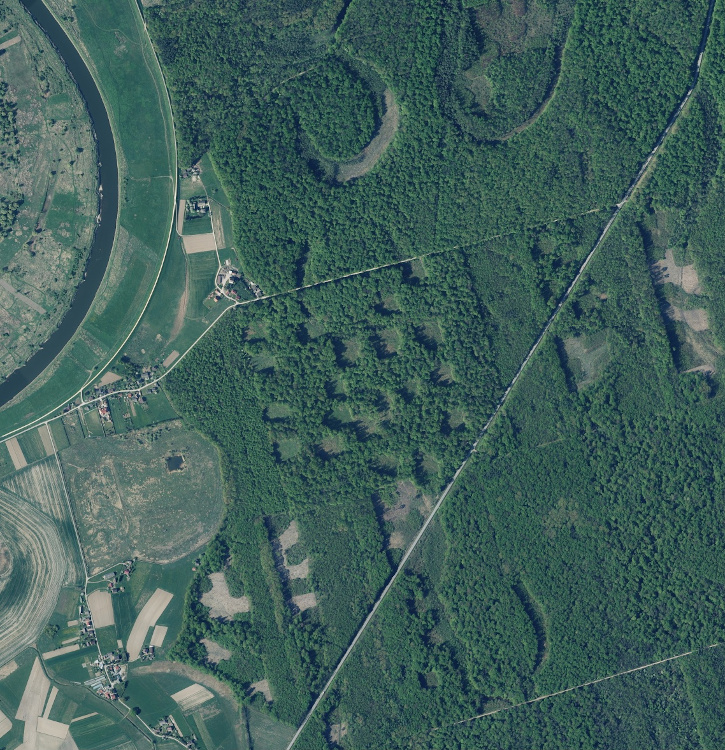}} 
    \subfigure{\includegraphics[width=0.49\linewidth, height=0.49\linewidth]{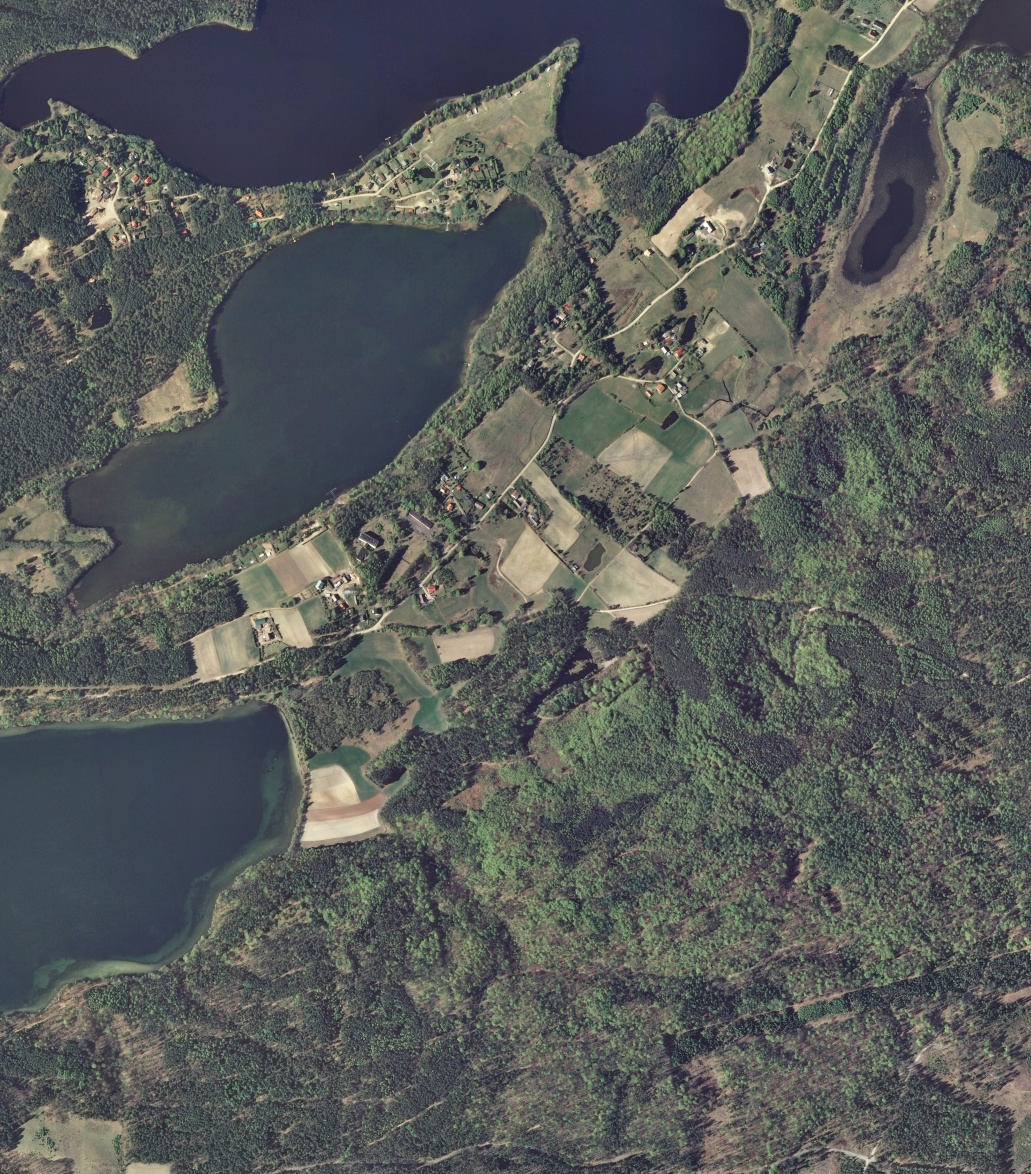}} 
    \subfigure{\includegraphics[width=0.49\linewidth, height=0.49\linewidth]{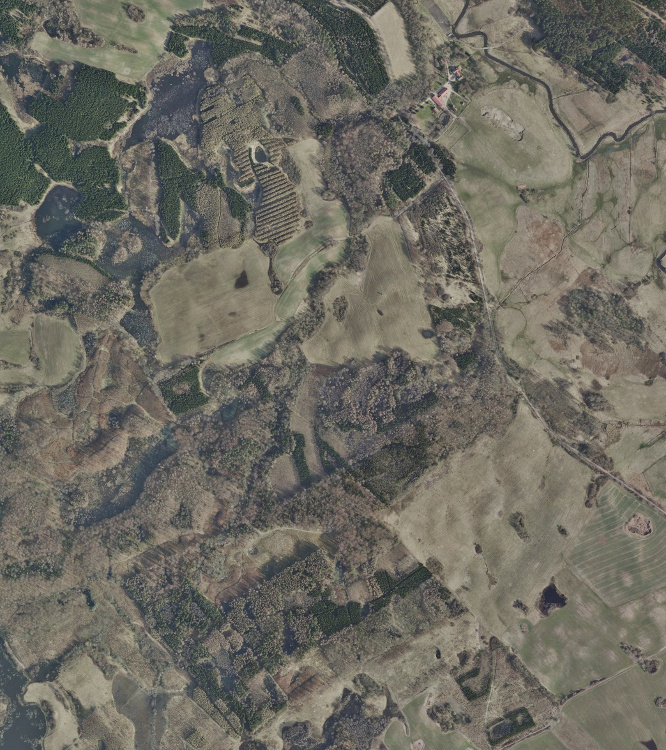}} \subfigure{\includegraphics[width=0.49\linewidth, height=0.49\linewidth]{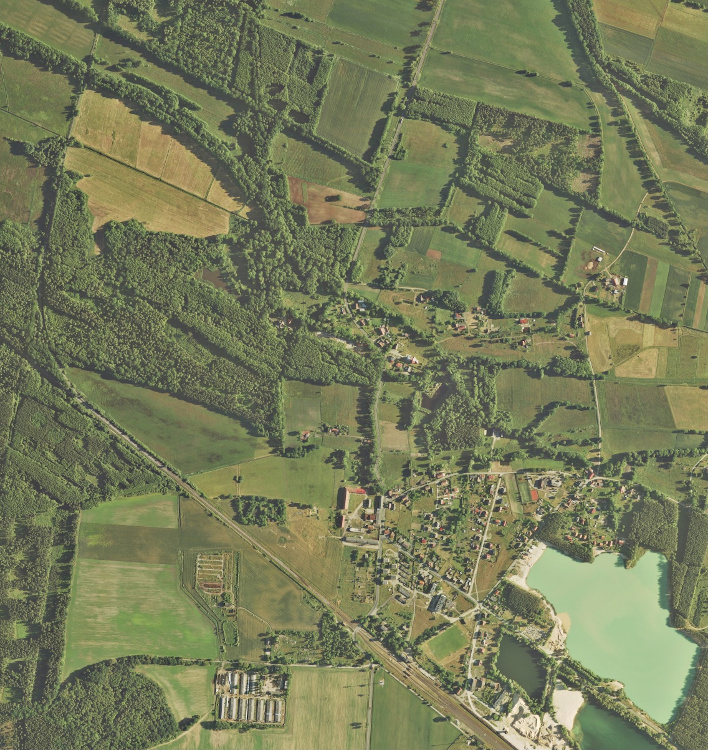}}
    \caption{Diversity of selected images. Different regions, seasons, time of day, weather, lighting conditions, etc.}
    \label{fig:samples}
\end{figure}

For the sake of maximum diversity of the dataset, we manually selected 41 orthophoto tiles from different counties located in all regions (as shown in Figure~\ref{fig:poland}). Every tile has about 5 km\textsuperscript{2}. There are 33 images with resolution 25~cm (ca. $9000\times9500$~px) and 8~images with resolution 50~cm (ca. $4200\times4700$~px), what gives 176.76 km\textsuperscript{2} and 39.51 km\textsuperscript{2} respectively and 216.27 km\textsuperscript{2} overall. Figure~\ref{fig:samples} shows samples of chosen images.

\subsection{Land cover characteristics}

The selected areas are located in Poland, i.e., in Central Europe. The majority of Poland spreads in the eastern part of the North European Plain. The country's geographic regions extend latitudinally, gradually passing from the lowlands in the north and center to highlands and mountains in the south of the country. The landscape is dominated by agricultural areas with a varied agrarian structure (60\%) as well as coniferous, deciduous, and mixed forests (29.6\%). The Polish forest cover is similar to the average of European (excluding Russia) and North American (both about 33\% of the area). Due to favorable climatic conditions, Poland's dominant forest type is coniferous forest (68.4\%, where pine accounts for 58\%). There are 38 urban agglomerations with more than 100 000 inhabitants, including one that exceeds 1 million. The extensive postglacial lake districts occupy the north of Poland, but numerous pounds are also scattered in the rest of the country.

\subsection{Classes}

We decided to annotate the images using four classes: building~(1), woodland~(2), water~(3), and road~(4) due to their usefulness and importance for public administration cases.

\begin{figure}[t]
    \centering
    \subfigure{\includegraphics[width=0.7\linewidth]{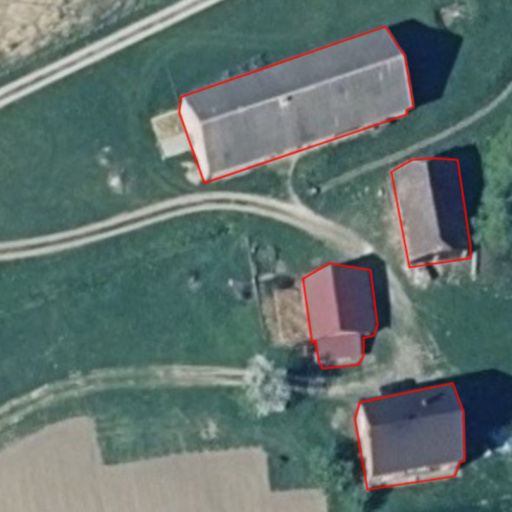}} 
    \caption{Class "building" means roof and all visible walls.}
\label{fig:building}
\end{figure}

\textbf{Building.}
An object standing permanently in one place. Greenhouses are excluded. Our images are not true orthophotos, so each building is annotated as roof and visible walls as shown in Figure~\ref{fig:building}.

\textbf{Woodland.}
Land covered with trees standing in proximity. Single trees and orchards are excluded.

\textbf{Water.}
Flowing and stagnant water including ponds and pools. Ditches and dry riverbeds are excluded.

\textbf{Road.}
The infrastructure used for road transport including parking and unpaved roads, and rail transport including tracks.

\textbf{Background.}
Area not classified to any class. It can include \eg fields, grass, pavements, and all objects excluded from above.

\begin{figure}[t]
    \centering
    \subfigure{\includegraphics[width=0.7\linewidth]{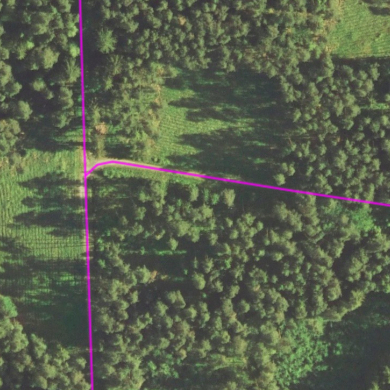}} 
    \caption{Narrow roads are annotated using polylines.}
\label{fig:road}
\end{figure}

\subsection{Annotations}

\begin{figure*}[t]
    \centering
    \subfigure{\includegraphics[width=0.195\linewidth]{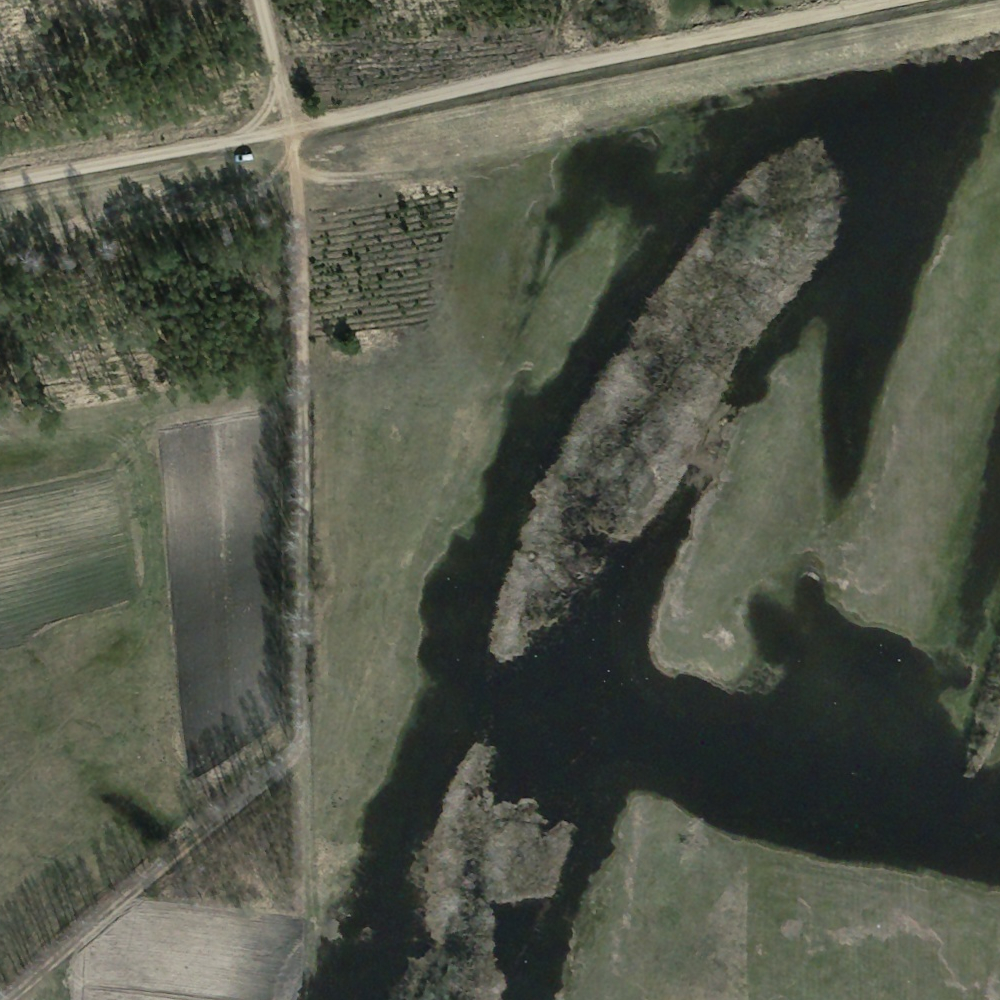}} 
    \subfigure{\includegraphics[width=0.195\linewidth]{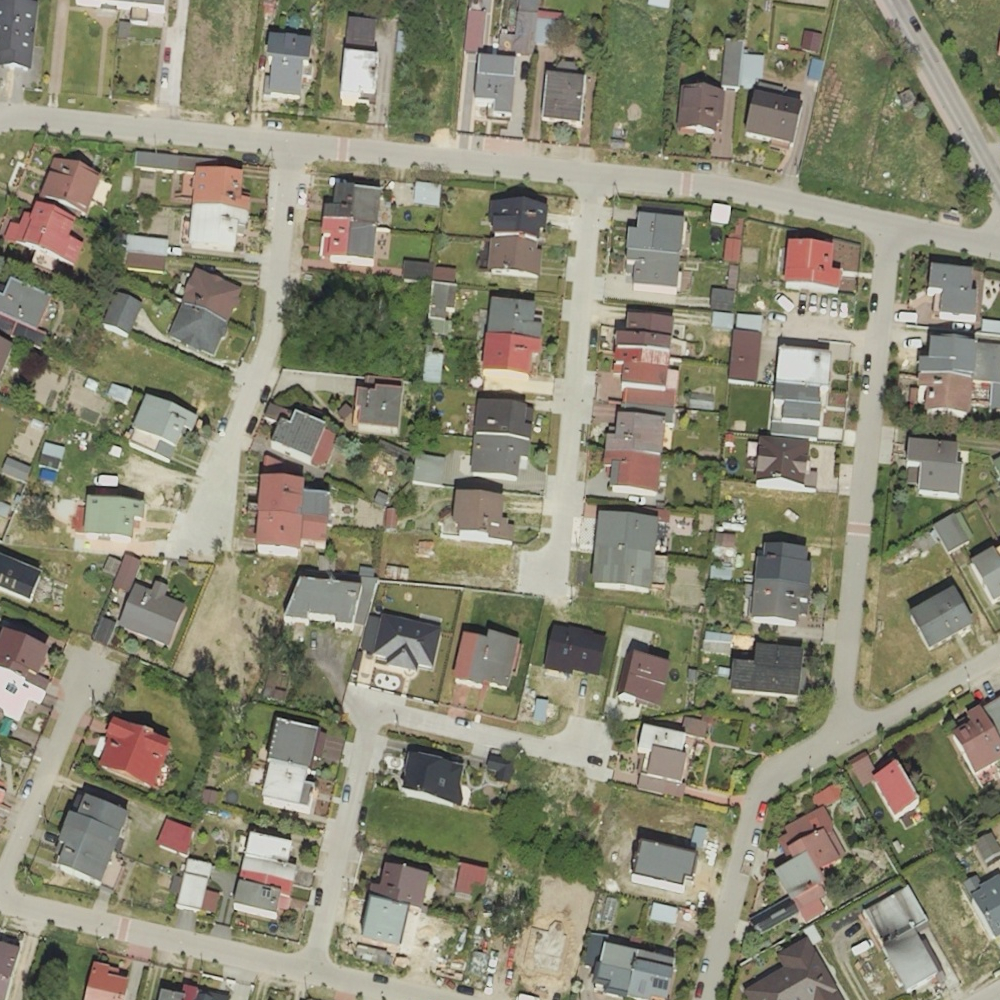}} 
    \subfigure{\includegraphics[width=0.195\linewidth]{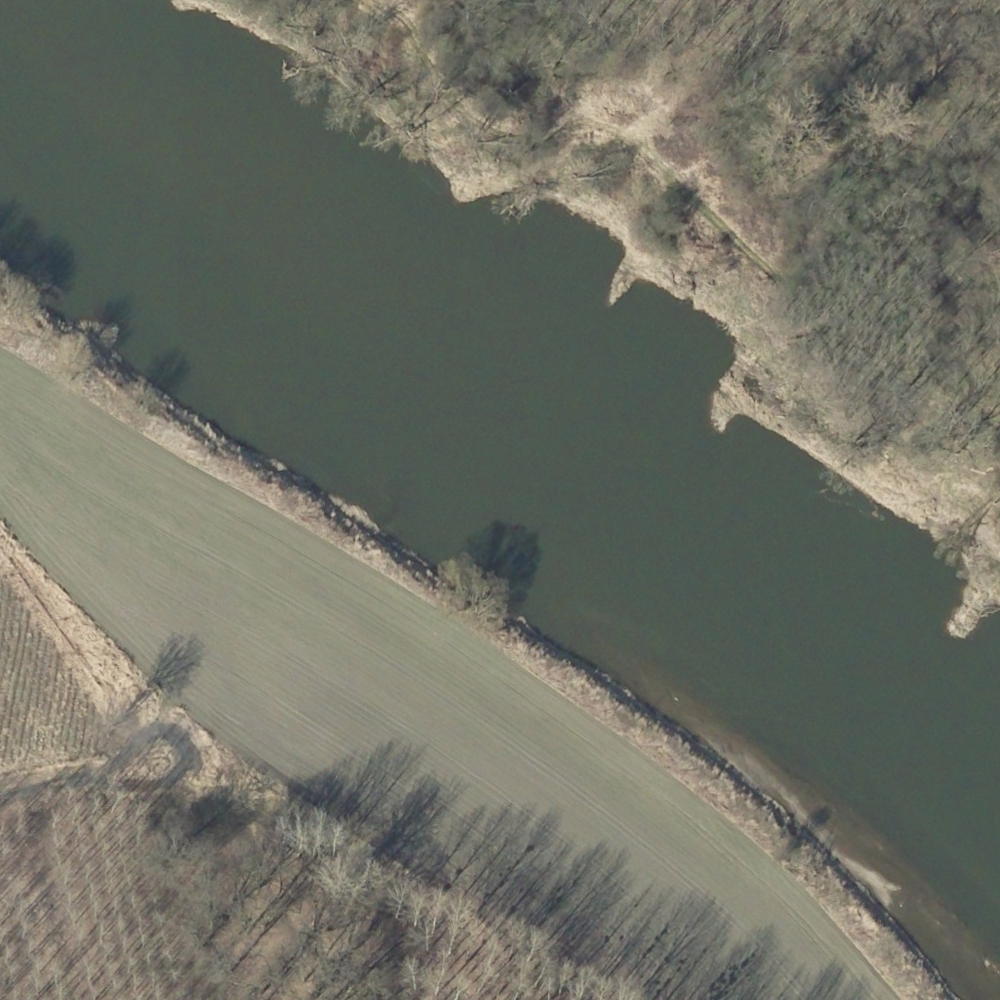}} 
    \subfigure{\includegraphics[width=0.195\linewidth]{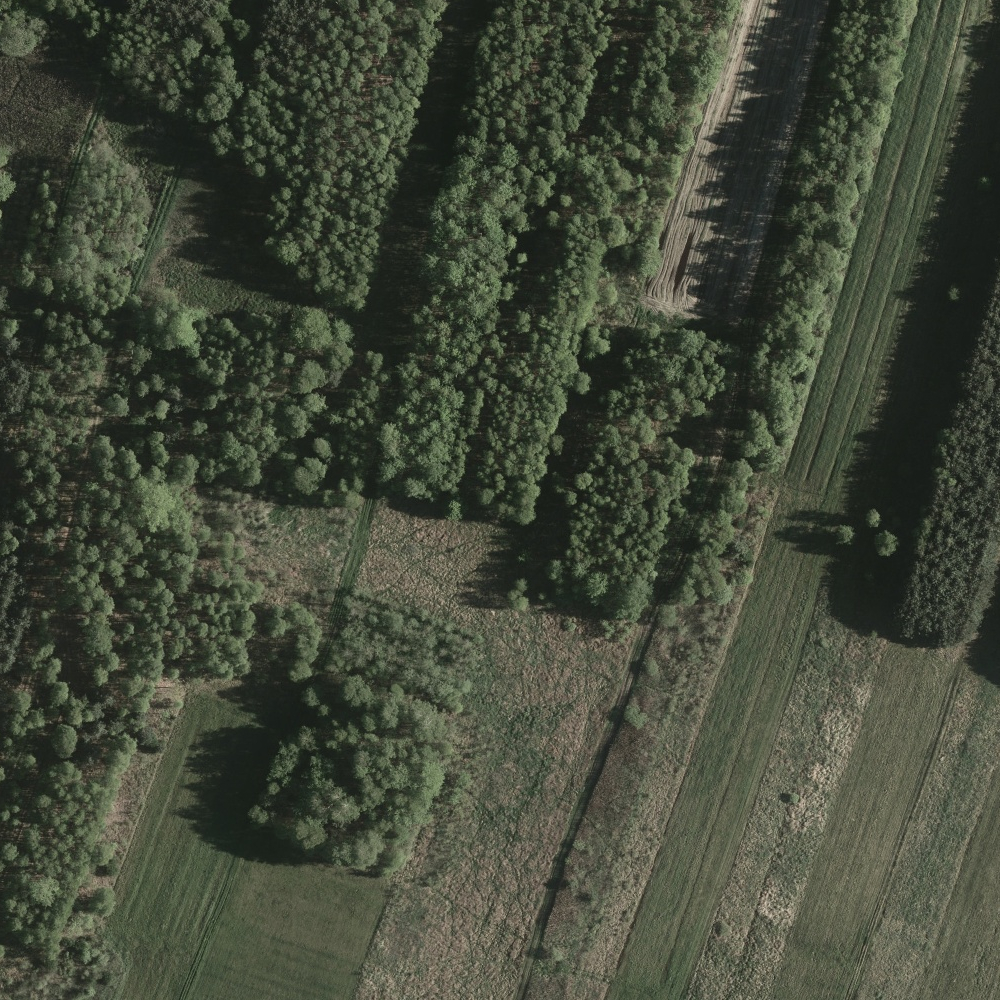}} 
    \subfigure{\includegraphics[width=0.195\linewidth]{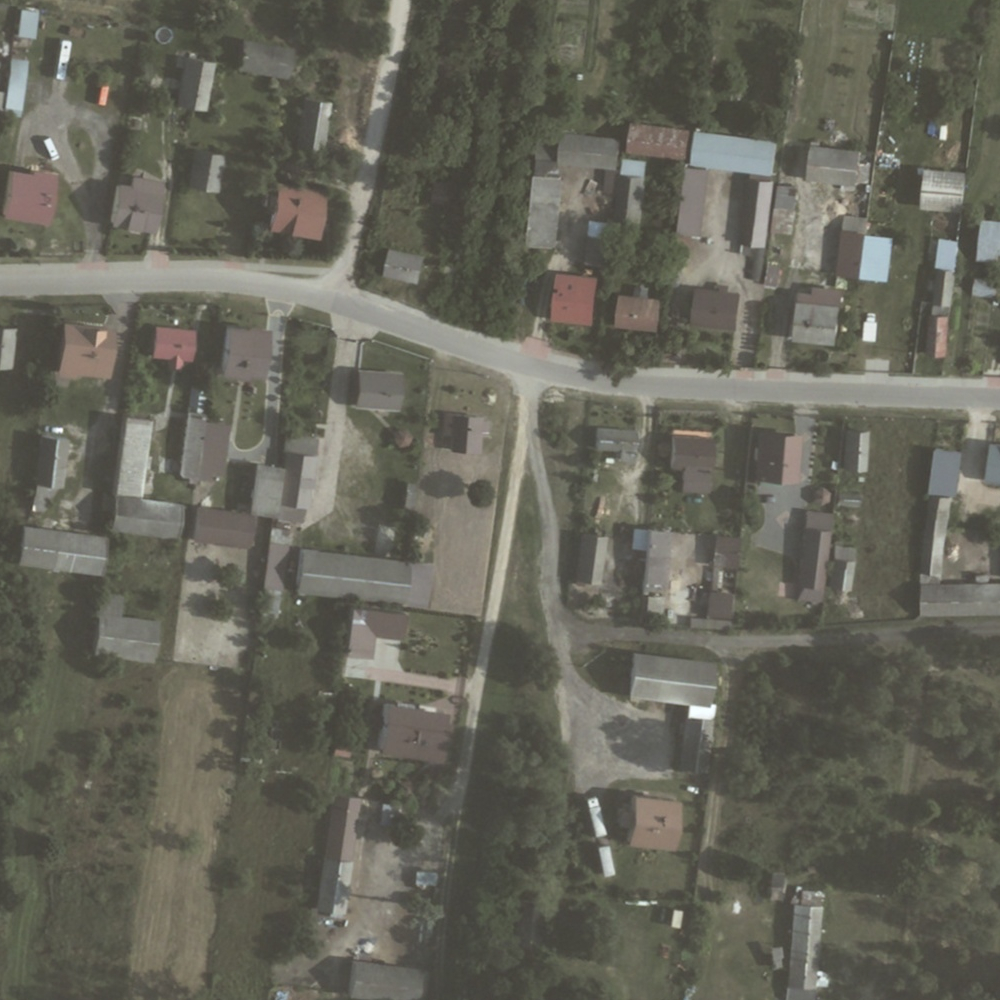}} 
    \newline
    \subfigure{\includegraphics[width=0.195\linewidth]{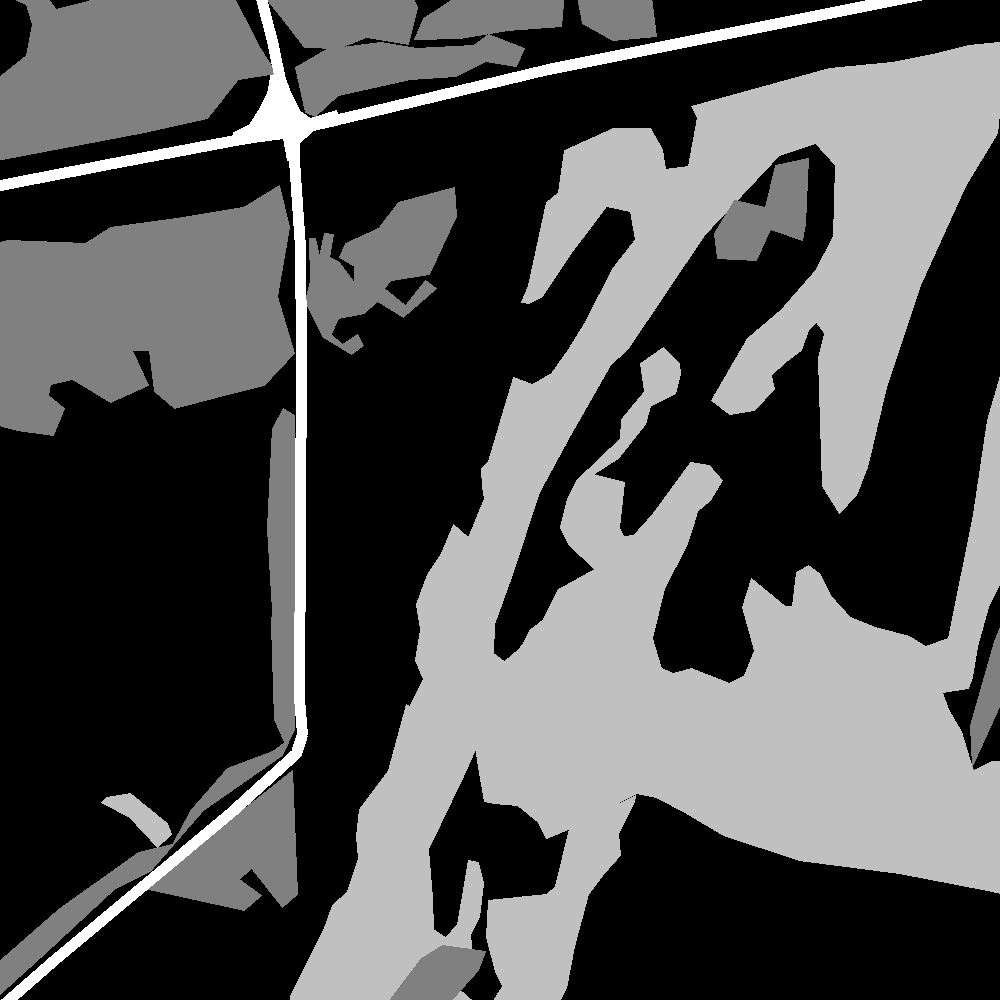}} 
    \subfigure{\includegraphics[width=0.195\linewidth]{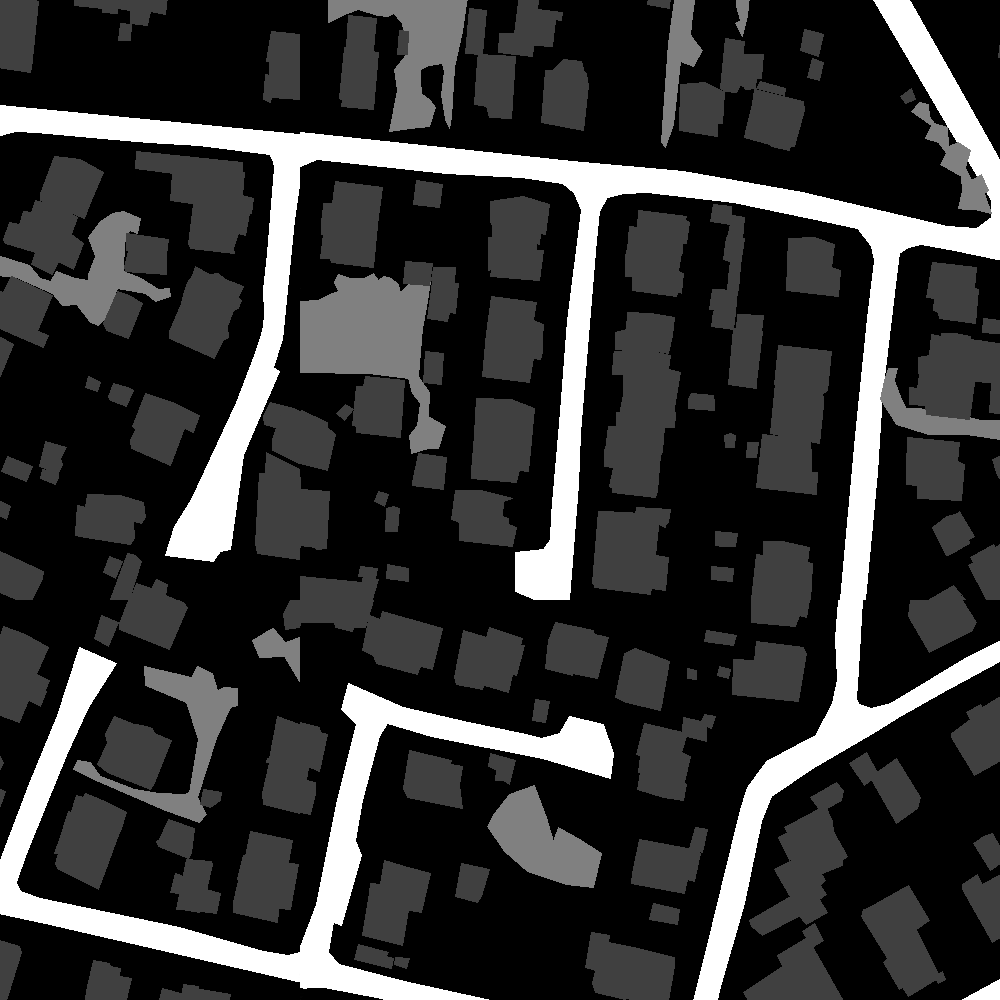}} 
    \subfigure{\includegraphics[width=0.195\linewidth]{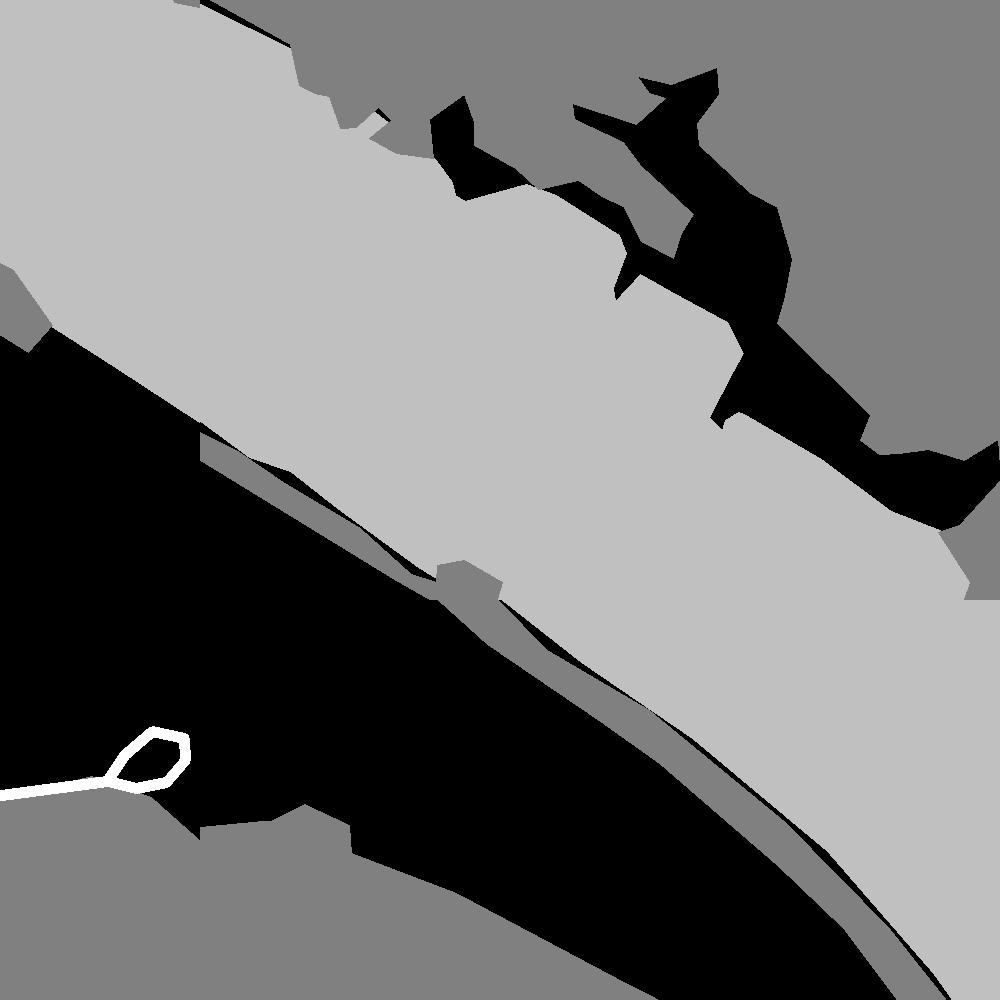}} 
    \subfigure{\includegraphics[width=0.195\linewidth]{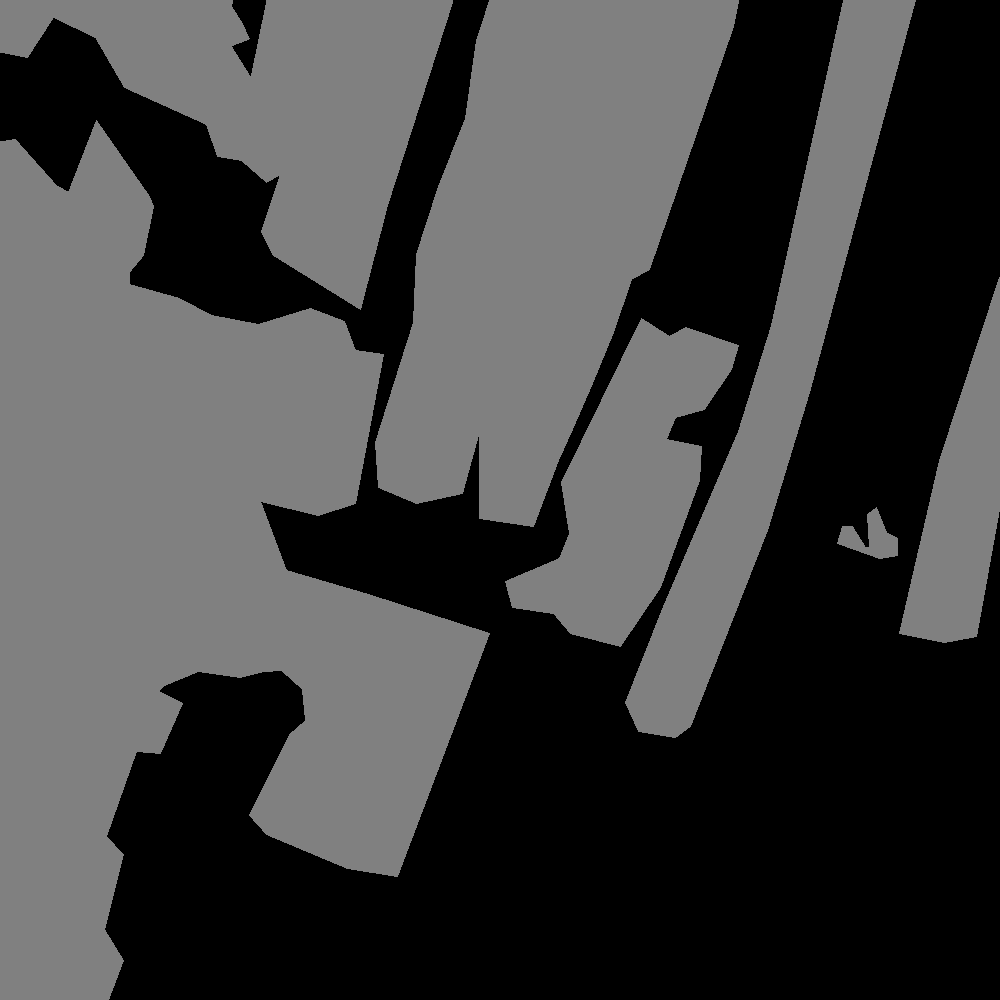}}
    \subfigure{\includegraphics[width=0.195\linewidth]{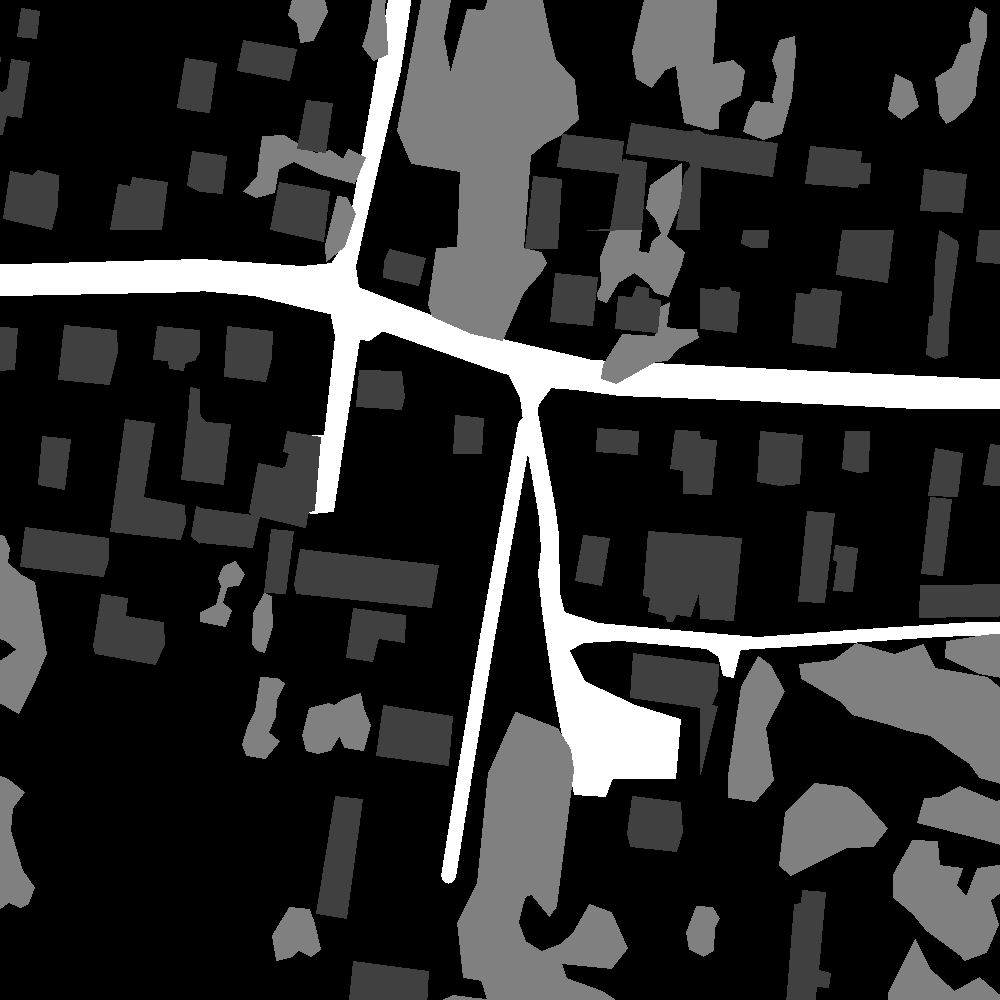}} 
    \caption{Close-ups of the images and their corresponding reference masks. Buildings are dark gray, woodlands are middle gray, water is light gray and roads are white.}
    \label{fig:mask}
\end{figure*}

Annotations are made manually with VGG Image Annotator (VIA) \cite{dutta2019vgg} by a group of people using polygon shape and polyline (only for narrow roads) as shown in Figure~\ref{fig:road}. Firstly we split every image into $2500\times2500$~px tiles for convenience. Tiles do not overlap, so the last tile in every row and all tiles in the last row are a little smaller. To provide a high-quality dataset, we implemented a rigorous procedure, so annotations are rather fine. A second person reviewed every finished tile. After that, all results were merged, and the segmentation mask was generated for each image, as shown in Figure~\ref{fig:mask}. The road mask was created by replacing a polyline with a polygon shape with fixed thickness.

Statistics are as follows. There are 12280 buildings (1.85~km\textsuperscript{2}), 72.02~km\textsuperscript{2} of woodlands, 13.15~km\textsuperscript{2} of water, 3.5~km\textsuperscript{2} of roads and 125.75~km\textsuperscript{2} of background in total.

\subsection{Comparison to related datasets}

Table~\ref{table:comparison} presents the comparison of the statistics between the proposed dataset and closely related aerial datasets: Inria \cite{maggiori2017dataset}, AIRS \cite{Chen_2019}, Massachusetts \cite{MnihThesis}, Agriculture-Vision \cite{chiu2020agriculturevision}, Tree Cover \cite{treecover}, ISPRS Potsdam and Vaihingen \cite{isprs}, and Chesapeake Conservancy \cite{robinson2019large}. The older ones have a worse resolution (Tree Cover, Massachusetts). Even newer high-resolution datasets usually cover one class \eg buildings (Inria, AIRS). ISPRS Potsdam and Vaihingen are high resolution but cover an urban area only. Agriculture-Vision and Chesapeake Conservancy were not available at the time of our dataset creation. Furthermore, they do not have building class. Despite Chesapeake is really large, it has automatic labels, which are coarse. Moreover, none of them except Potsdam and Vaihingen are located in Europe, where land cover (\eg forest type) can be different. 

In summary, there is no dataset with particular classes (buildings, trees, water, roads) covering a rural area with a resolution of tens centimeters per pixel and manual fine labels.

\begin{table*}[t]
\begin{center}
\begin{tabular}{m{0.18\linewidth}<{\centering}|m{0.18\linewidth}<{\centering}|m{0.18\linewidth}<{\centering}|m{0.09\linewidth}<{\centering}|m{0.09\linewidth}<{\centering}|m{0.13\linewidth}<{\centering}}
\hline\noalign{\smallskip}
Dataset & Location & Classes & Coverage (km\textsuperscript{2}) & Resolution (cm/px) & Annotations\\
\noalign{\smallskip}
\hline\hline
\noalign{\smallskip}
Inria & USA/Austria & buildings & 810 & 30 & semi-automatic \\
AIRS & Christchurch (New~Zealand) &  buildings & 457 & 7.5 & semi-automatic \\
Massachusetts Buildings & Boston (USA) &  buildings & 340 & 100 & automatic \\
Massachusetts Roads & Boston (USA) & roads & 2600 & 100 & automatic \\
Tree Cover & São~Paulo (Brazil) & trees & 8000 & 100 & automatic \\
Agriculture-Vision & USA & 9 (excluding buildings and trees) & $\sim$560 & 10/15/20 & manual \\
ISPRS Potsdam & Potsdam (Germany) & 5 (excluding water) & $\sim$1.4 & 5 & manual \\
ISPRS Vaihingen & Vaihingen (Germany) & 5 (excluding water) & $\sim$3.4 & 9 & manual \\
Chesapeake Conservancy & Chesapeake Bay (USA) & 13 (excluding buildings) & $\sim$160,000 & 100 & automatic \\
\noalign{\smallskip}
\hline
\noalign{\smallskip}
\textbf{\datasetname (ours)} & \textbf{Poland} & \textbf{buildings, woodlands, water, roads} & \textbf{216}  & \textbf{25/50}  & \textbf{manual}\\
\noalign{\smallskip}
\hline
\end{tabular}
\end{center}
\caption{Comparison of similar aerial datasets for semantic segmentation. All of them have RGB channels except Agriculture-Vision and ISPRS Potsdam and Vaihingen, which additionally have a near-infrared (NIR) band. Chesapeake Conservancy has 6 infrared channels also. Further, Agriculture-Vision, ISPRS Potsdam and Vaihingen coverage are estimated, as they do not provide this information.}
\label{table:comparison}
\end{table*}

\section{Experiments}

In order to know how general semantic segmentation networks perform on our dataset and to check if \datasetname can be useful, we decided to create a baseline model. We chose one of state-of-the-art architectures - DeepLabv3+ \cite{deeplabv3plus2018} using modified Xception71 \cite{deeplabv3plus2018,xception,deformable} with Dense Prediction Cell (DPC) \cite{dpc2018} as a backbone. Additionally, we performed a few more experiments to check if data augmentation and some model modifications can improve results and, if so, how much. 

\subsection{Data preparation}

Firstly we split 41 images and their corresponding masks into $512\times512$ tiles, getting rid of smaller ones (these on the right and bottom edges), and we shuffled them. Then we organized it as follows: 15\% (1602) of tiles became test set, 15\% (1602) of tiles became validation set, and the last 70\% (7470) became train set. We provide lists of the filenames with the dataset.

\begin{figure}[t]
    \centering
    \subfigure{\includegraphics[width=0.325\linewidth]{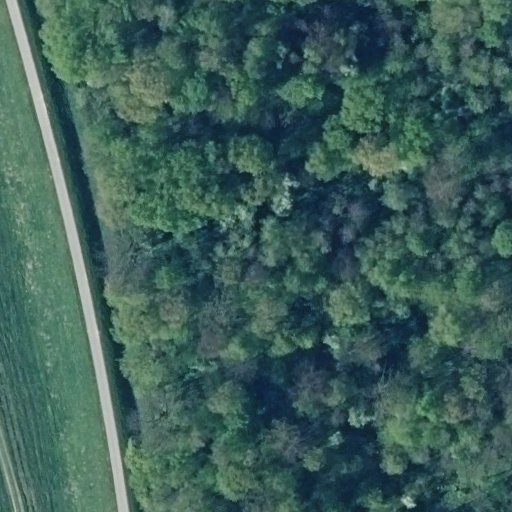}}  \subfigure{\includegraphics[width=0.325\linewidth]{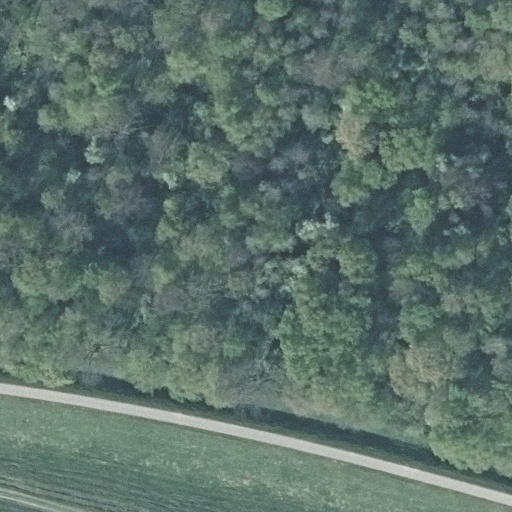}} 
    \subfigure{\includegraphics[width=0.325\linewidth]{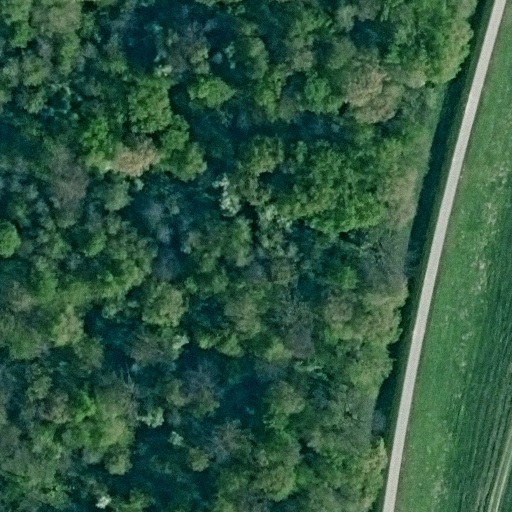}} 
    \newline
    \subfigure{\includegraphics[width=0.325\linewidth]{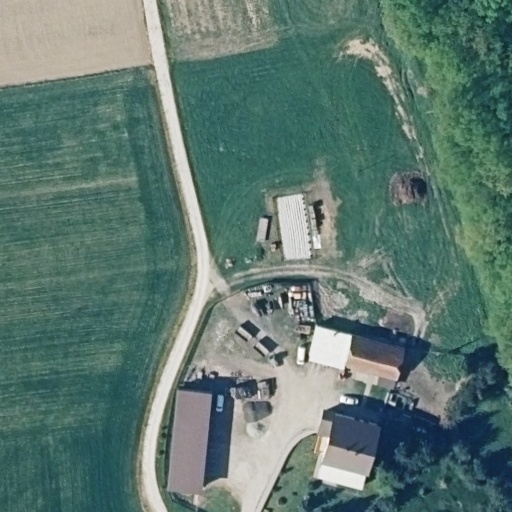}}
    \subfigure{\includegraphics[width=0.325\linewidth]{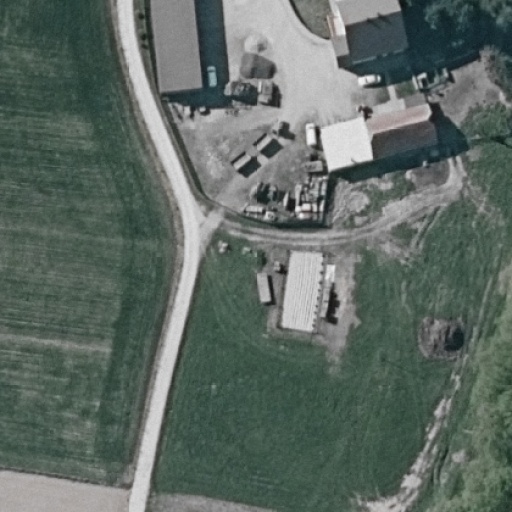}} 
    \subfigure{\includegraphics[width=0.325\linewidth]{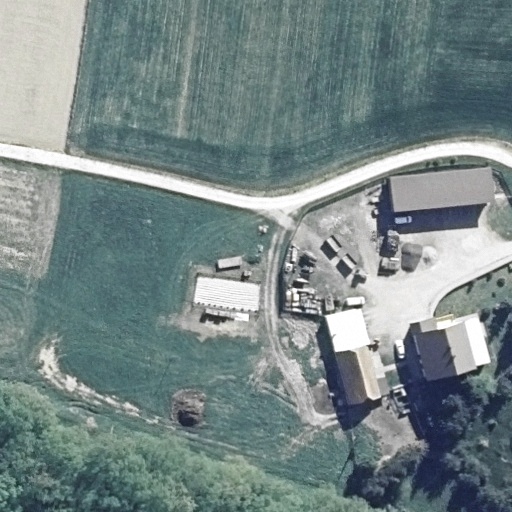}} 
    \newline
    \subfigure{\includegraphics[width=0.325\linewidth]{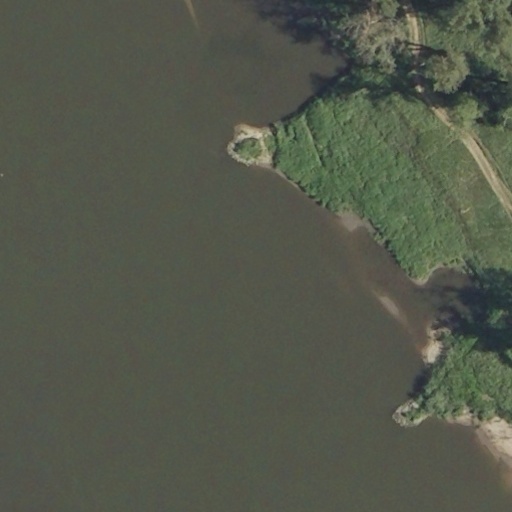}}  
    \subfigure{\includegraphics[width=0.325\linewidth]{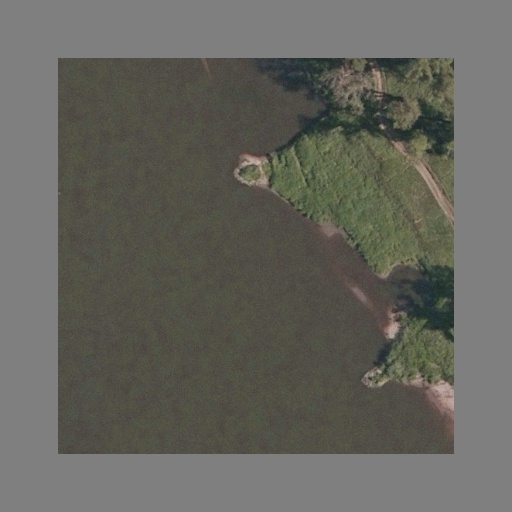}} 
    \subfigure{\includegraphics[width=0.325\linewidth]{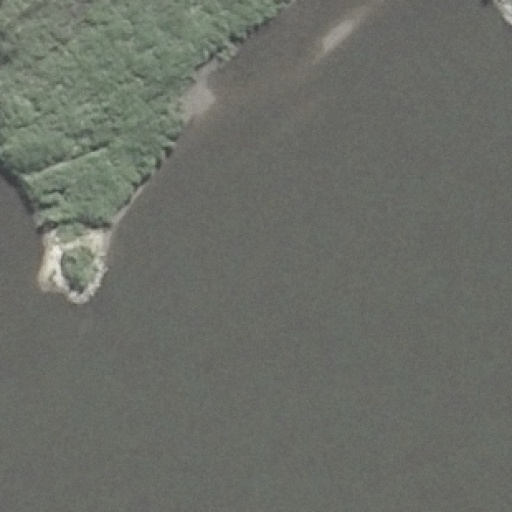}} 
    \caption{First sample in every row is the original image. The next, augmented images try to simulate various: seasons, lighting conditions, haziness, sizes of objects, etc.}
    \label{fig:augmentation}
\end{figure}

\subsection{Data augmentation}

We believe that proper augmentation, which simulates different flights and diverse land cover conditions, could be useful, so we applied an offline augmentation to the training set with imgaug \cite{imgaug}. We added nine augmented copies for every tile, randomly changing the following parameters: hue, saturation, grayscale, contrast, brightness, sharpness, adding noise, doing flipping, rotation, cropping, and padding. Therefore, we reached 74700 samples in the training set. Figure~\ref{fig:augmentation} presents sample augmentations.

\subsection{Training}

To train the network, we utilized a single NVIDIA Titan RTX GPU with 24GB of memory. We used Xception71 DPC pretrained on Cityscapes \cite{Cordts_2016_CVPR} to initialize weights and set decoder output stride = 4. Then we performed a few experiments changing encoder output stride from 16 to 4 along with batch size and atrous rates accordingly. We provided appropriate loss weights to counteract unbalanced area sizes of particular classes also. The other hyperparameters remained as default (as described in \cite{deeplabv3plus2018}). 

To evaluate the model, we use the mean intersection over union (mIoU), which is the standard metric for semantic segmentation. The mIoU is the average of intersection over union (IoU) across all classes. And IoU is defined as the area of overlap between the ground truth and predicted class divided by the area of their union.

We finished the training when there was no significant gain of mIoU on the validation set.

\subsection{Results}

\begin{figure*}[t]
    \centering
    \subfigure{\includegraphics[width=0.195\linewidth]{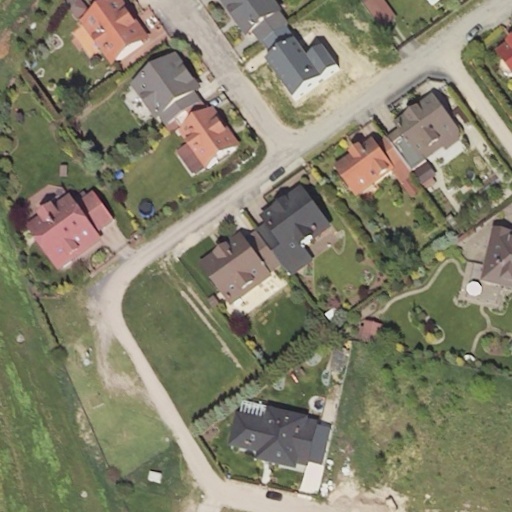}}     \subfigure{\includegraphics[width=0.195\linewidth]{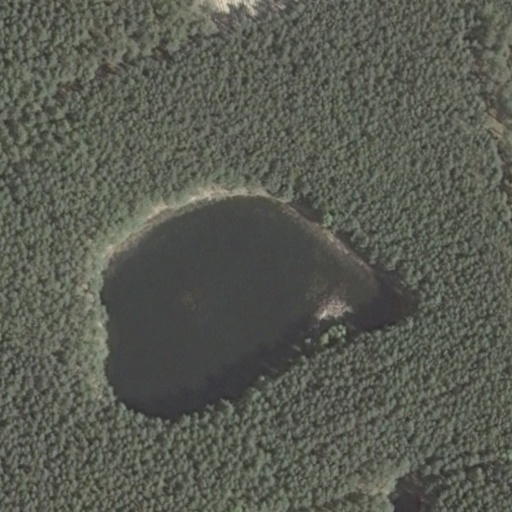}} 
    \subfigure{\includegraphics[width=0.195\linewidth]{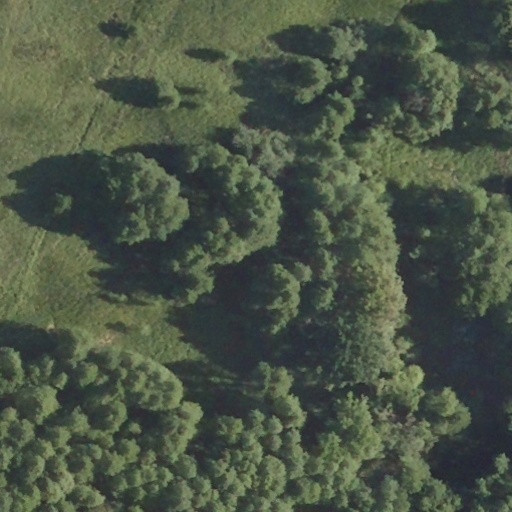}} 
    \subfigure{\includegraphics[width=0.195\linewidth]{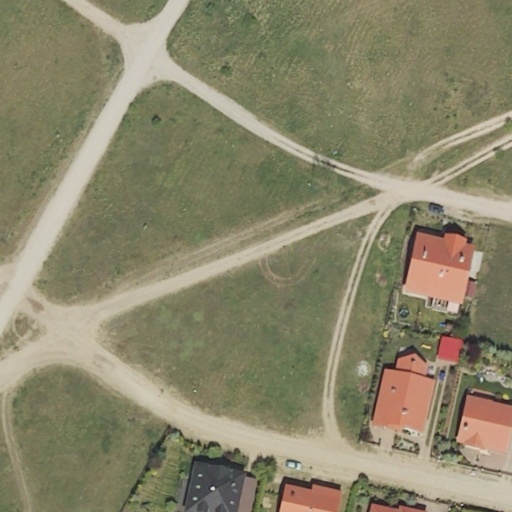}} 
    \subfigure{\includegraphics[width=0.195\linewidth]{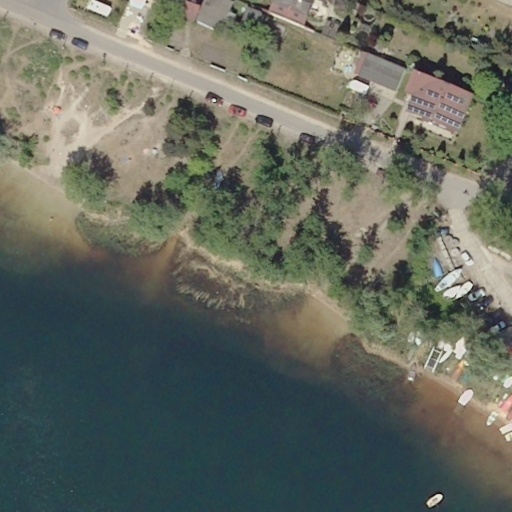}} 
    \newline
    \subfigure{\includegraphics[width=0.195\linewidth]{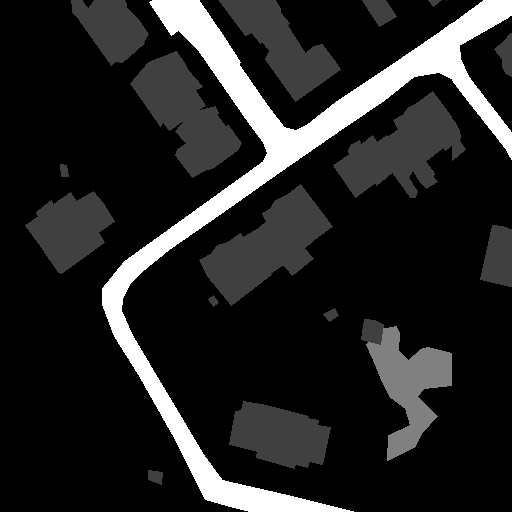}}     \subfigure{\includegraphics[width=0.195\linewidth]{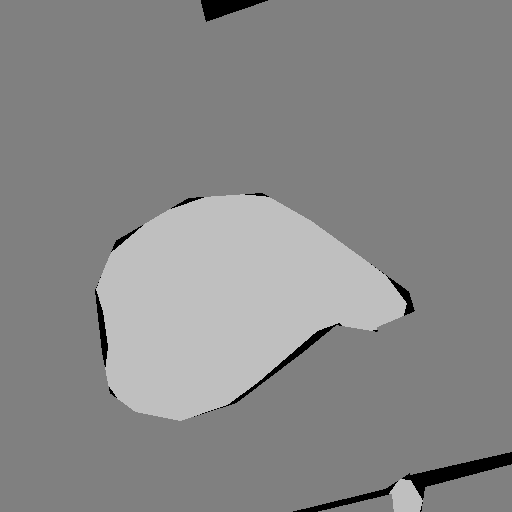}} 
    \subfigure{\includegraphics[width=0.195\linewidth]{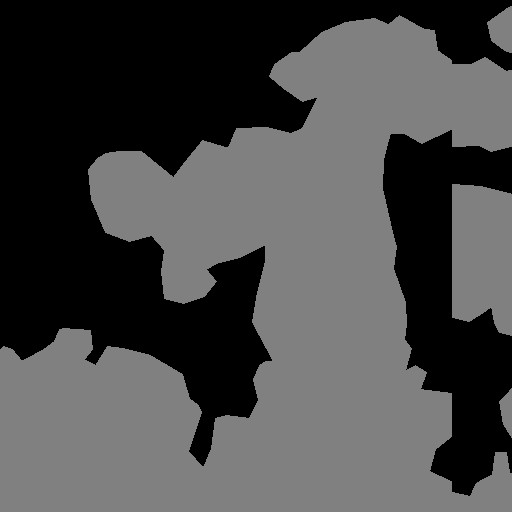}} 
    \subfigure{\includegraphics[width=0.195\linewidth]{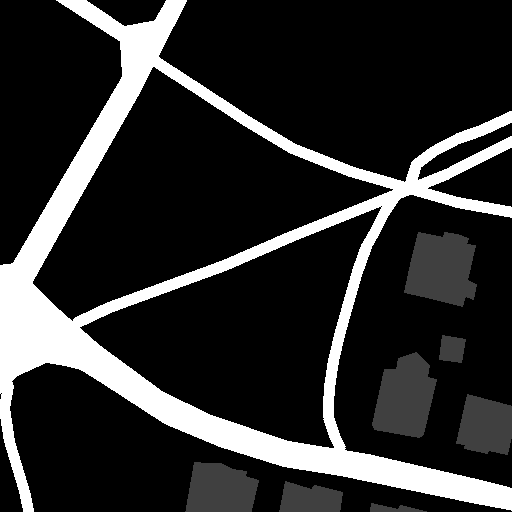}}
    \subfigure{\includegraphics[width=0.195\linewidth]{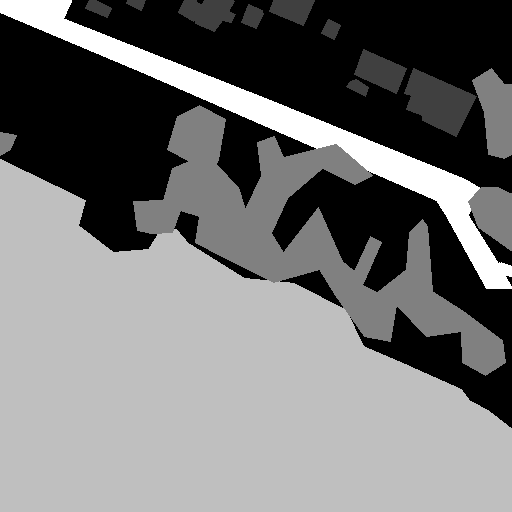}} 
    \newline
    \subfigure{\includegraphics[width=0.195\linewidth]{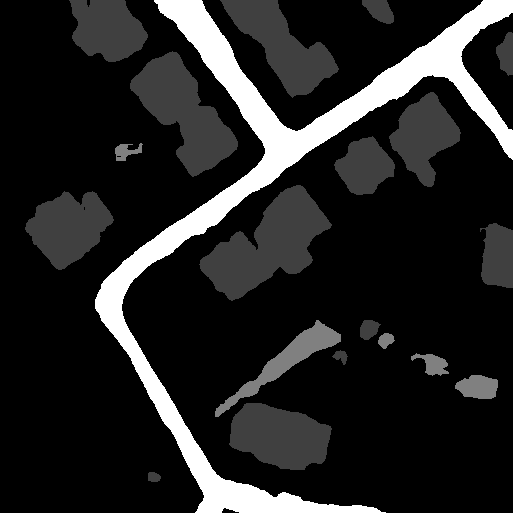}} 
    \subfigure{\includegraphics[width=0.195\linewidth]{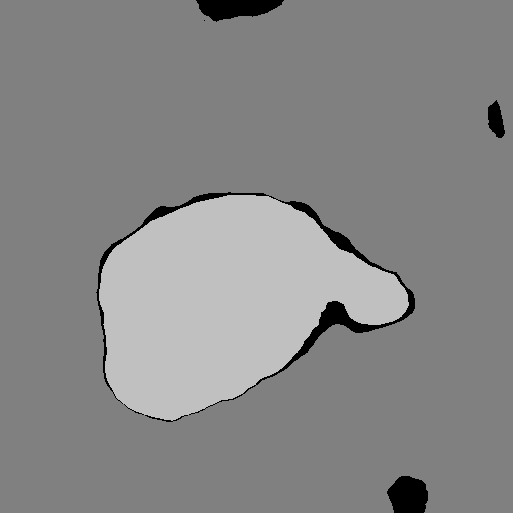}}     \subfigure{\includegraphics[width=0.195\linewidth]{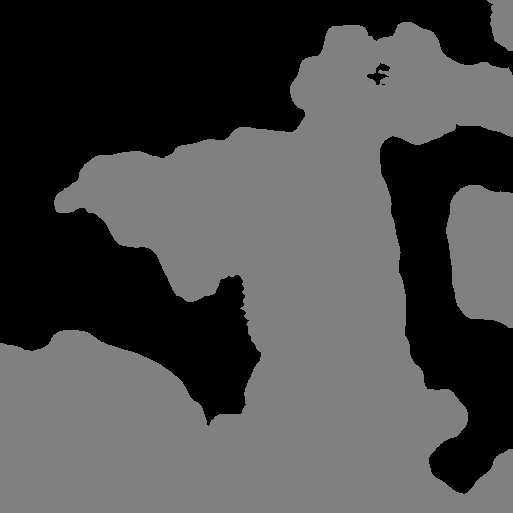}} 
    \subfigure{\includegraphics[width=0.195\linewidth]{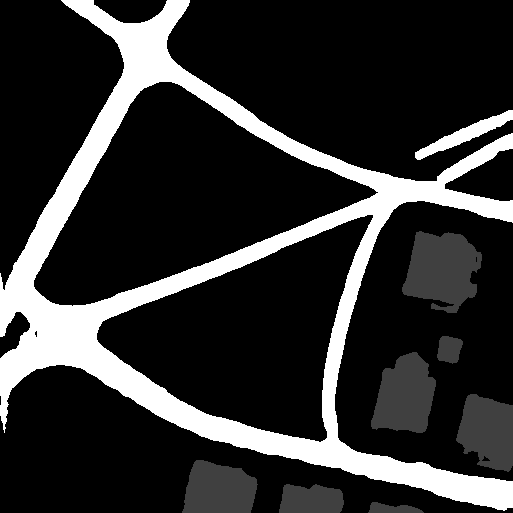}} 
    \subfigure{\includegraphics[width=0.195\linewidth]{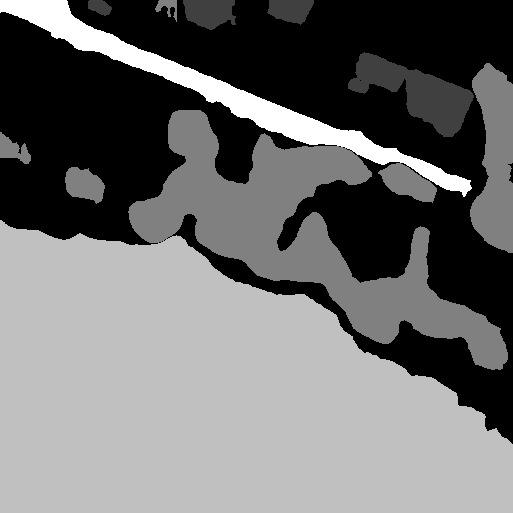}}
    \caption{Close-ups of the images (first row), their corresponding annotations (second row), and experiment results (third row). Detected areas usually have much smoother edges, which better fits reality in the case of woodlands and water, but is some imprecise in the case of buildings. Some small buildings are omitted (first from the right) and roads have slightly jagged edges (first from the left), on the other hand, a neural network can find trees we missed (first from the right).}
    \label{fig:results}
\end{figure*}

In table \ref{table:results}, we report the results obtained on the test set using IoU metrics. Figure~\ref{fig:results} shows close-ups of images, labels, and result segmentation. 

Roads and buildings are the most challenging classes for semantic segmentation, as they are often narrow (roads) or small (buildings). Hence they have fewer inner pixels, which are easier to classify correctly. In that case, imprecise edges cause greater error. Moreover, they are sometimes obscured by other objects like trees.

The baseline DeepLabv3+ model reaches 81.81\% of mIoU of the entire test set. Smaller output stride provides better results, finally giving 84.09\%. Augmentation further improves the metrics by 1.47\% reaching 85.56\%.

The results prove that automatic mapping from aerial images is possible with deep learning and a relatively small dataset.

\begin{table*}[t]
\begin{center}
\begin{tabular}{p{0.2\linewidth}<{\raggedright}|p{0.1\linewidth}<{\centering}|p{0.1\linewidth}<{\centering}|p{0.1\linewidth}<{\centering}|p{0.1\linewidth}<{\centering}|p{0.1\linewidth}<{\centering}|p{0.1\linewidth}<{\centering}}
\hline\noalign{\smallskip}
Method & Buildings & Woodlands & Water & Roads & Background & Overall \\
\noalign{\smallskip}
\hline\hline
\noalign{\smallskip}
DeepLabv3+ OS 16 & 74.12\% & 89.89\% & 93.01\% & 59.96\% & 92.05\% & 81.81\% \\
DeepLabv3+ OS 8 & 77.47\% & 90.62\% & 93.77\% & 62.64\% & 92.65\% & 83.43\% \\
DeepLabv3+ OS 4 & 77.53\% & 91.05\% & 93.84\% & 65.04\% & 93.02\% & 84.09\% \\
\noalign{\smallskip}
\hline
\noalign{\smallskip}
\textbf{DeepLabv3+ OS 4 +~augmentation} & \textbf{79.74\%} & \textbf{91.46\%} & \textbf{94.39\%} & \textbf{68.74\%} & \textbf{93.45\%} & \textbf{85.56\%} \\
\noalign{\smallskip}
\hline
\end{tabular}
\end{center}
\caption{Intersection over union on the test set and the mean of all classes (Overall). OS denotes encoder output stride during training and evaluation. The smaller output stride, the better results. We proved data augmentation is useful and improves the results in every class.}
\label{table:results}
\end{table*}

\section{Conclusions}

In this work, we present a unique RGB-only \datasetname dataset with aerial data typical for Central Europe, manually annotated for four classes: buildings, woodlands, water, and roads. The dataset is high resolution (tens centimeters per pixel), covers various rural areas and contain images with various optical conditions and periods of the vegetation season. In order to prove the usefulness of the dataset, we provide the results of a few baseline experiments using a state-of-the-art deep learning model - DeepLabv3+.

As we demonstrate, the dataset can be used to create tools for automatic mapping using neural networks. This allows for improving the efficiency and accuracy of identifying changes in land use and land cover. Therefore, it can be beneficial in various domains, such as administration, agriculture, forestry, and water resource management. Moreover, \datasetname fills an important gap as there was a lack of open aerial datasets useful for this type of application. 

In the future, we plan to develop \datasetname by adding more classes \eg fields, ditches, as well as splitting existing general classes like water into detailed \eg lake, river, pond, and pool. 

We make this dataset publicly available to encourage its future development and use.

{\small
\bibliographystyle{ieee_fullname}
\bibliography{egbib}
}

\end{document}